\documentclass{article}
\usepackage{spconf,amsmath,graphicx}

\title{BATINet: Background-Aware Text to Image Synthesis and Manipulation Network}
\name{Ryugo Morita, Zhiqiang Zhang, Jinjia Zhou}
\address{Faculty of Science and Engineering, Hosei University, Tokyo, Japan \\
ryugo.morita.7f@stu.hosei.ac.jp}
\begin{document}
%\ninept
%
\maketitle
\begin{abstract} 
Background-Induced Text2Image (BIT2I) aims to generate foreground content according to the text on the given background image. 
Most studies focus on generating high-quality foreground content, although they ignore the relationship between the two contents. 
In this study, we analyzed a novel Background-Aware Text2Image (BAT2I) task in which the generated content matches the input background. We proposed a Background-Aware Text to Image synthesis and manipulation Network (BATINet), which contains two key components: Position Detect Network (PDN) and Harmonize Network (HN). The PDN detects the most plausible position of the text-relevant object in the background image. The HN harmonizes the generated content referring to background style information. Finally, we reconstructed the generation network, which consists of the multi-GAN and attention module to match more user preferences. Moreover, we can apply BATINet to text-guided image manipulation. It solves the most challenging task of manipulating the shape of an object. We demonstrated through qualitative and quantitative evaluations on the CUB dataset that the proposed model outperforms other state-of-the-art methods.
\end{abstract}
\section{Introduction}
\label{sec:intro}
% 現状の紹介
Many researchers in recent years have proposed a lot of methods based on GAN \cite{tao2022df}\cite{ye2022recurrent}\cite{zhu2019dm} and diffusion models \cite{saharia2022photorealistic}\cite{ramesh2021zero}\cite{nichol2021glide} in Text2Image field.
It has been regarded as having a high potential for many applications such as education, image editing software, and video game development. However, a few opportunities exist to generate images from scratch for practical use. Even with such an opportunity, it would randomly generate the background content to match the foreground content generated according to the text, which is not flexible.
% 私が研究に興味を持った理由
From those perspectives, the Background-Induced Text2Image (BIT2I) piqued our interest.

With the rapid development of GAN \cite{goodfellow2014generative} technology, the BIT2I has achieved excellent results. MC-GAN \cite{park2018mc} generated a decent foreground image according to the text. Moreover, BGNet \cite{wang2021background} generated fidelity images. However, both methods generate not only foreground content but also text-irrelevant content (e.g., branches, ground, etc.) to make the image more realistic.
In other words, although the BIT2I does not generate foreground content according to the text while matching the given background, it generates foreground content according to the text on the given background. Therefore, we defined the Background-Aware Text2Image (BAT2I) in which the generated image matches the input background image.
% 以降で既存研究の問題

The existing model for this task has three problems:
(1) content generation in a random position without attention to the background.
(2) inconsistent background and generated foreground style owing to separate image processing, and 
(3) scope for improvement in image quality and fidelity.
% それを解決するために
To address these problems, we proposed a Background-Aware Text to Image synthesize Network (BATINet), which can detect the position and generate the foreground content according to the text while matching the background. It consists of three networks: Position Detect Network (PDN), Generation Network(GN), and Harmonization Network (HN).
% モデルの紹介を軽くする
The general flow of the model is as follows. First, the PDN detects the most plausible position of the text-relevant object in the background image. 
Then, the GN generates the foreground content according to the text and position information obtained by the PDN. Finally, we adapt the HN before the last generator to be consistent between these generated content and original background content styles.
Moreover, we can apply BATINet to text-guided image manipulation. It solves the most challenging task of manipulating the shape of an object.

% 実験結果はどうなったのか
Extensive experiments on the CUB datasets demonstrate that our proposed method can generate high-quality images that match the background.
We demonstrated through qualitative and quantitative evaluations that the proposed model outperforms other state-of-the-art methods.

% figure
\begin{figure*}[htbp]
\centerline{\includegraphics[width=\linewidth]{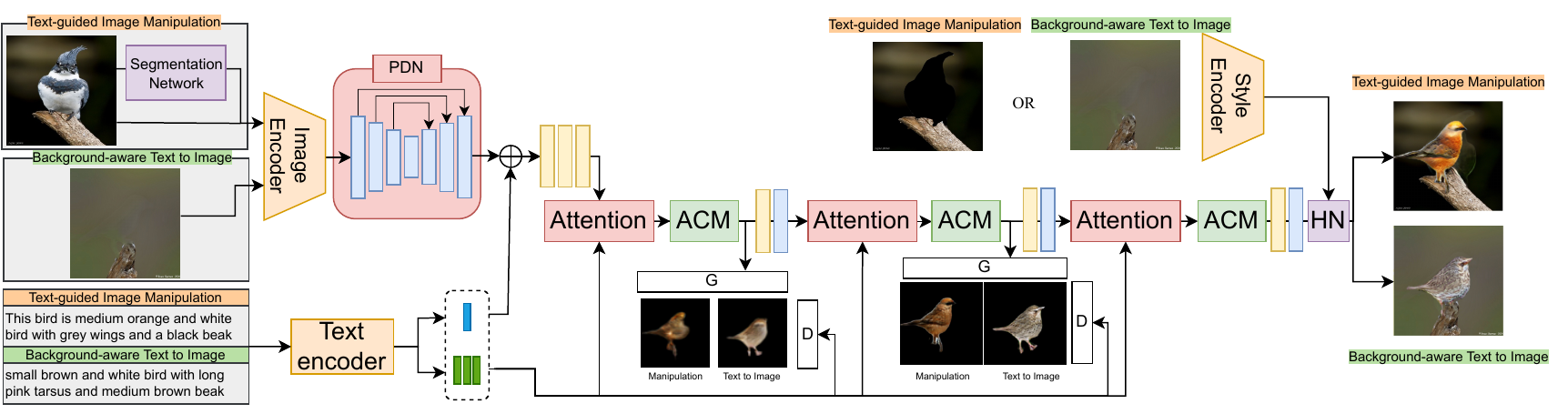}}
\caption{Overview of our proposed model. For the BAT2I, the inputs are the background image and a text description. The background image detects the most likely location of the foreground image. For the TGIM, the inputs are the image and a text description. The image detects the location of the object that exists to output position information. In addition, a foreground image is generated according to the text and position information. Finally, the style information of the generated foreground and background images is harmonized to generate a more realistic image.}
\label{model}
\end{figure*}

\section{Method}
\label{sec:method}
Fig. \ref{model} shows the architecture of the proposed model. For the BAT2I, the inputs are a given background image $I_{b}$ and a text description $S$. 
The target is to synthesize the foreground content $I_{f}'$ according to the text $S$ while considering the background $I_{b}$.
First, the PDN is applied to detect the plausible position to output as a position vector $P_{obj}$. 
Second, we reconstruct the Generation Network to synthesize foreground image $I_{f}$ according to the text $S$ and position vector $P_{obj}$.
Finally, the output image $I'$ is obtained by the HN, which inputs the generated foreground content $I_{f}$ and the given original image $I_{b}$. 
Its network harmonizes the difference between the styles of the two contents.
For the TGIM, the inputs are a normal image $I_{n}$ and a text description $S$.
First, the PDN is applied to detect the object's position to output as a position vector $P_{obj}$. 
The subsequent process is performed according to the BAT2I task.
Our model consists of three networks: PDN, GN, and HN. Next, we introduce the specific content of each architecture.

\begin{figure}[htbp]
\centerline{\includegraphics[width=0.9\linewidth]{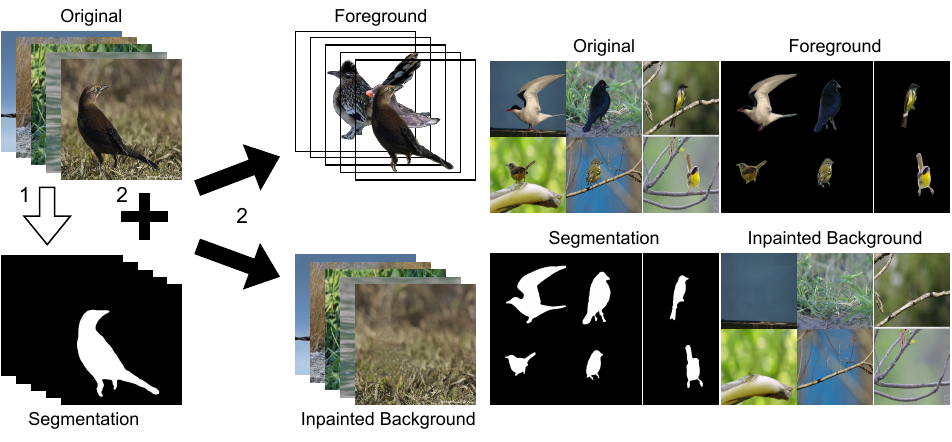}}
\caption{By using pretrained models, we create provisional datasets; foreground, segmentation, and inpainted background images.}
\label{data}
\end{figure}

\subsection{Dataset}
The existing datasets cannot efficiently perform the BAT2I or TGIM tasks. Therefore, we will handle them by creating provisional datasets with the pretrained models.
Based on the CUB dataset \cite{wah2011caltech}, we created three extra datasets: a segmentation mask of an object, an object-only foreground image, and an inpainted background image dataset, as shown in Fig. \ref{data}.
First, we collected the segmentation mask dataset of an object using deeplabv3\cite{chen2018encoder}, which is trained to acquire segmentation masks for birds. The segmentation rendered the object and non-object areas 1 and 0, respectively.
Second, we masked the original image with the segmentation mask obtained from the previous process to collect the object-only foreground image dataset.
Finally, we inpainted the original dataset with segmentation masks using pretrained HiFill\cite{yi2020contextual} to collect an inpainted background image dataset.

\begin{figure*}[htbp]
\centerline{\includegraphics[width=0.88\linewidth]{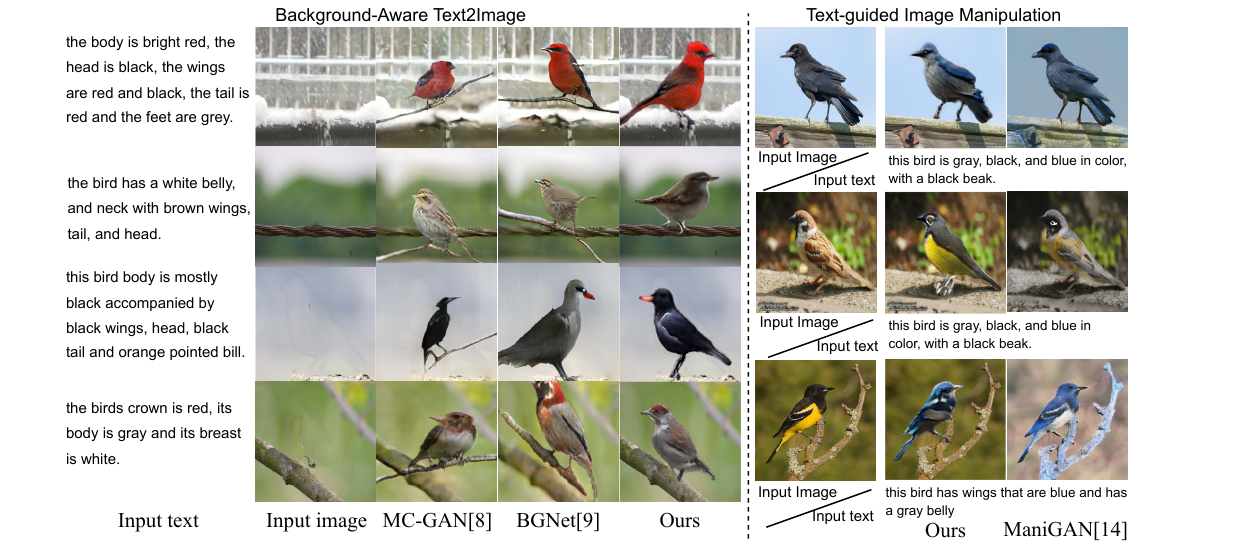}}
\caption{Qualitative comparison of the proposed method and other existing methods. For the BAT2I, the existing model generates images by adding text-irrelevant content such as a branch for image reality. By contrast, the proposed method detects the plausible location of the object in the background to generate the foreground content. For the TGIM, existing methods can only manipulate style information. However, the proposed method manipulates not only style information but also the object's shape according to the text. Furthermore, the background image is well-preserved.}
\label{result}
\end{figure*}

\subsection{Position Detect Network (PDN)}
Existing models generate text-relevant foreground content without considering the background content. 
Therefore, as shown in Fig. \ref{result}, they often generate branches or foundations in the background image to render it more realistic. 
To address this issue, we proposed the PDN which detects the most plausible position of the text-relevant object in the background image to provide its information to the GN. 
Specifically, we adapted a naive image style transfer model \cite{isola2017image} to detect the location for generating foreground content.
Additionally, we successfully obtain the position information by using the intermediate feature of the image transfer model that changes the background into the object's segmentation mask. 
Therefore, the PDN generator trains to input the background $I_{b}$ or normal $I_{n}$ image and output the object position vector $P_{obj}$. Its P enables us to decide the object's position that best matches the background image. We describe the detail of the training in section \ref{sec:train}.

\subsection{Generation Network (GN)}
% 既存研究はどのようなものか
Inspired by the BGNet \cite{wang2021background}, we also introduced the Multi-GAN for this task. 
% 既存研究の問題点
However, BGNet does not fully leverage the benefits of Multi-GAN for image quality and fidelity.
% それを解決するために...
Therefore, we reconstructed a novel GN, which consists of the HN, the attention module, and the tr-ACM network based on ManiGAN \cite{li2020manigan}.
% それがどのようなものであるか説明
The model can generate the object, enabling the image to align with the input text description and the position vector obtained by the PDN. 
Therefore, the position vector is the first input to the model from which it should generate the object. The model can generate foreground content according to the text description while satisfying the requirement for the position.
% 最後に具体的な(input, output, modelの中身の説明を行う)
To associate image-text representations, we adopted a text-relevant affine combination module (tr-ACM) used in \cite{Morita_2023_WACV}. It accepts the fused features after attention and the position features obtained by the PDN. Because ManiGAN's ACM processes feature for the entire image, the contents of regions unrelated to the text were also retained as text-image representation. In contrast, the tr-ACM is a fine-grain text-image representation that combines only text and only text-relevant regions.

\subsection{Harmonization Network (HN)}
% 既存研究はどのようなものか
Most existing methods do not consider the gap in style information between generated foreground and background images, even if they were processed separately.
% それによって発生する問題
It leads to cases in which a dark light background image would produce brightly styled foreground content and vice versa.
% それを解決するために...
To address this issue, we proposed HN.
% それがどのようなものであるか説明
Specifically, with pretrained Harmonizer \cite{ke2022harmonizer}, the style information of the generated foreground content can be harmonized with the style information from the given background image to ensure higher consistency.

% \begin{figure}[htbp]
% \centerline{\includegraphics[width=\linewidth]{img/ICIP-result2.pdf}}
% \caption{現実の写真.}
% \label{result2}
% \end{figure}

\subsection{Loss function and training}
\label{sec:train}
Except for the HN, the other two networks are trainable; because they can be trained from scratch, they are also helpful for more specific professional image synthesis.
In the training process, the entire configuration is based on GAN: it contains one and three GANs for the PDN and the GN, respectively. In addition, the GN has three tr-ACMs. The PD and the GN are trained separately.

\textbf{Position Detect Network.} According to \cite{isola2017image}, the generator loss $L_{G_{pdn}}$ for the PDN consists of an adversarial loss $L_{adv}$ and an L1 loss $L_{L1}$. We define it as follows:
\begin{eqnarray}
    L_{L1} = E_{x,y,z} [||y - G(x,z)||]\\
    L_{G_{pdn}} = L_{adv} + L_{L1}
\end{eqnarray}
The loss function of the discriminator only includes the adversarial loss, and the specific equation is as follows:
\begin{eqnarray}
    L_{D} = L_{adv}
\end{eqnarray}

\textbf{Generation Network.}
According to \cite{li2020manigan}\cite{Morita_2023_WACV}, the generator loss $L_{G}$ consists of an adversarial loss $L_{adv}$, a perceptual loss $L_{per}$, a text-image matching loss $L_{DAMSM}$, and a regularization loss $L_{reg}$.  The definition can be expressed as follows:
\begin{equation}
  L_{reg} = 1-\frac{1}{CHW}||I^{'}_{tr}-I_{tr}|| 
\end{equation}
\begin{equation}
\fontsize{8pt}{0cm}\selectfont
\begin{split}
    L_{G} = L_{adv} + L_{per} + {1 - L_{cor}(I_{tr}^{'}, S)} \\ + L_{DAMSM}(I', S) + L_{reg},
\end{split}
\end{equation}
where $C$, $H$, and $W$ are the number of color channels, the height, and width of the input image $I_{tr}$, respectively.
The loss function of the discriminator only includes the adversarial loss, and the specific equation is as follows:
\begin{equation}
\fontsize{8pt}{0cm}\selectfont
\begin{split}
L_{D} = L_{adv} + {1 - L_{cor}(I_{tr}^{'},S)} + L_{cor}(I_{tr}^{'},S^{'}),
\end{split}
\end{equation}
where $S^{'}$ is a randomly selected mismatched textual description from the dataset.

\section{Experiment}
\label{sec:experiment}
We evaluated methods using the CUB dataset \cite{wah2011caltech}, which contains 11,788 bird images of 200 categories. A total of 8,855 and 2,933 images were used for training and testing, respectively. We compared the proposed model with state-of-the-art models in qualitative and quantitative performance.
We trained PDN and GN with 600 epochs on the CUB dataset.

% Tableを生成
\begin{table}[tb]
  \caption{Quantitative comparison results between the proposed method and other existing methods.}
  \label{table:qualitative}
  \centering
  \setlength{\tabcolsep}{3mm}
  \resizebox{\columnwidth}{!}{%
  \begin{tabular}{l|ccccc}
    \hline
    \multicolumn{5}{c}{CUB\cite{wah2011caltech}}\\
    \hline
     & MC-GAN \cite{park2018mc} & BGNet \cite{wang2021background} & Ours w/o PDN & Ours w/o HN & Ours\\
    \hline
        IS $\uparrow$ & 2.90 & 3.00 & 3.41 & 3.43 & \textbf{3.45}\\
        NIMA $\uparrow$  & 5.42 & 5.61 & 5.62 & 5.61 & \textbf{5.63}\\
    \hline
  \end{tabular}}
\end{table}

\subsection{Quantitative comparison}
In the quantitative experiment, we evaluated the IS \cite{salimans2016improved} and NIMA \cite{talebi2018nima} on randomly selected images from the CUB dataset with a randomly selected text description. A total of 10,000 images were generated for quantitative evaluation. The results are listed in Table \ref{table:qualitative}. Our method outperforms state-of-the-art models in all metrics. In the IS, our generated image produces highly discriminative images. Furthermore, the NIMA of the proposed method is also high, indicating that the proposed method generates high-quality manipulated images. 

\subsection{Qualitative comparison}
% \textbf{On inpainted image dataset,} 
Fig. \ref{result} compares the existing models and the proposed method.
The BAT2I results indicate that the proposed model can consider the background for generating the foreground content according to the text description.
In the existing models, foreground images were generated in random positions or added perches, and the ground was generated to ensure realism.
By contrast, the proposed model can generate more realistic images by training to understand the position in the background image, which has the highest possibility to present the generated object.
Furthermore, the foreground content generated by existing models has limitations in resolution and fidelity. However, by creating the model, the proposed method addressed that part of the problem. In addition, by using the HN to solve the style gap caused by processing foreground and background images separately, we generated more realistic images.
We describe these advantages of each network in section \ref{sec:ablation}.

For the TGIM task, the PDN can detect the object's position, which needs to be manipulated in the input normal image.
Furthermore, the GN generates an object according to the text at the same position as the original object to achieve the TGIM task.
The TGIM result shows that the proposed model can modify not only the object’s style information (e.g., texture, color, etc.) but also the form of the object.

\begin{figure}[htbp]
\centerline{\includegraphics[width=0.83\linewidth]{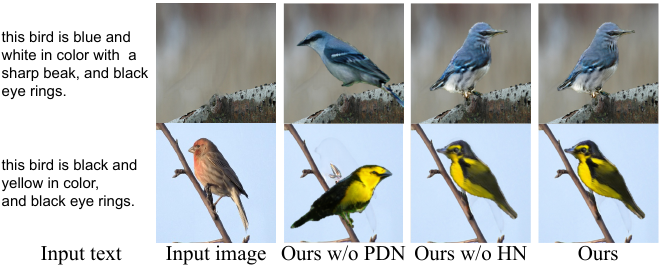}}
\caption{Ablation studies of the proposed method and without PDN and without HN. The PDN enables the model to detect the high possible position to exist. The HN enables to harmonize the style of background and foreground content.}
\label{ablation_result}
\end{figure}

\subsection{Ablation study}
\label{sec:ablation}
To better understand the benefit of the proposed network: the PDN and the HN, we show the comparison result without each network in Fig. \ref{ablation_result} and Table. \ref{table:qualitative}.
The model without the PDN generates the bird in a strange position, whereas the model with the PDN generates the bird to be on perch or on the ground. It implies that it helps consider the best position in the background.
In the model without the HN, the style of generated content and the given background image has a gap, which leads to output discomfort. By contrast, the model with the HN enavles it to harmonize its gap. It indicates that it helps the output image to be more realistic. In the quantitative results, models with each network model outperform those without them in all metrics.

\section{Conclusion}
We proposed BATINet, which can detect the most appropriate location and generate the foreground object according to the text while matching the input background image. We introduced the PDN that detects the most plausible position of the text-relevant object in the background image, and the HN that harmonizes the generated content referring to background style information.
In addition, the proposed method not only manipulated style but formed information in text-guided image manipulation tasks.
We demonstrated through qualitative and quantitative evaluations on the CUB dataset that the proposed model outperforms other state-of-the-art methods.

\bibliographystyle{IEEEbib}
\bibliography{refs}

\end{document}